# Physics-Based Explainable AI for ECG Segmentation: A Lightweight Model

Muhammad Fathur Rohman Sidiq
Department of Physics, Faculty of Mathematics and Science, Brawijaya University, Malang, Indonesia
Abdurrouf
Department of Physics, Faculty of Mathematics and Science, Brawijaya University, Malang, Indonesia
Didik Rahadi Santoso *
Department of Physics, Faculty of Mathematics and Science, Brawijaya University, Malang, Indonesia

* Corresponding author. E-mail: dieks@ub.ac.id

## Abstract

The heart's electrical activity, recorded through Electrocardiography (ECG), is essential for diagnosing various cardiovascular conditions. However, many existing ECG segmentation models rely on complex, multi-layered architectures such as BiLSTM, which are computationally intensive and inefficient. This study introduces a streamlined architecture that combines spectral analysis with probabilistic predictions for ECG signal segmentation. By replacing complex layers with simpler ones, the model effectively captures both temporal and spectral features of the P, QRS, and T waves. Additionally, an Explainable AI (XAI) approach is applied to enhance model interpretability by explaining how temporal and frequency-based features contribute to ECG segmentation. By incorporating principles from physics-based AI, this method provides a clear understanding of the decision-making process, ensuring reliability and transparency in ECG analysis. This approach achieves high segmentation accuracy—97.00% for the QRS wave, 93.33% for the T wave, and 96.07% for the P wave. These results indicate that the simplified architecture not only improves computational efficiency but also provides precise segmentation, making it a practical and effective solution for heart signal monitoring.

## 1 Introduction

Electrocardiography (ECG) is a fundamental tool for diagnosing and monitoring cardiovascular conditions by recording the heart's electrical activity [1–3]. The main components of the ECG waveform—namely the P-wave, QRS complex, and T-wave—carry crucial physiological information that reflects underlying cardiac function. Accurate segmentation of these waves is essential for identifying arrhythmias, ischemia, and other cardiac abnormalities. However, reliable segmentation remains challenging due to morphological variability, signal noise, and inter-patient differences [4–6].

Deep learning approaches, particularly Convolutional Neural Networks (CNNs) and Recurrent Neural Networks (RNNs), have been widely employed for ECG signal analysis [4–6]. Hybrid models such as CNN-BiLSTM have shown promising results by combining spatial and temporal modeling, thereby improving segmentation accuracy [7–10]. Nevertheless, many of these models function as black boxes, making it difficult to interpret how they differentiate between ECG components. Moreover, segmentation performance may degrade in the presence of overlapping waveforms or poorly delineated signals [9,11–14].

To address these limitations, this study introduces a physics-based Explainable AI (XAI) approach for ECG segmentation, integrating well-established signal processing techniques to enhance interpretability and performance [9,11–14]. Rather than depending solely on deep models for feature extraction, we incorporate mathematically grounded preprocessing transformations that emphasize physiologically meaningful signal features. These include:

1. Hilbert Transform – extracts the instantaneous amplitude and phase of the ECG signal, aiding in wave differentiation.

2. Euler Differentiation – enhances rapid changes in the signal, making it easier to detect sharp transitions such as QRS onset and offset.
3. Gauss-Legendre Integration – smooths the signal while preserving wave morphology, effectively expanding waveform structures to improve segment separation [5,15–18].

Additionally, Fast Fourier Transform (FFT) is employed to analyze the spectral characteristics of each preprocessing method, demonstrating how different transformations influence frequency-domain representations of ECG waves [18–22]. This spectral analysis provides further insights into the effectiveness of preprocessing in improving segmentation performance[23–25]. This study proposes a lightweight CNN-BiLSTM model that integrates these physics-based preprocessing techniques to achieve robust and interpretable ECG segmentation. The key contributions of this research include:

1. Physics-based preprocessing for ECG segmentation: The combination of Hilbert Transform, Euler Differentiation, and Gauss-Legendre Integration provides a structured mathematical framework for distinguishing ECG waves, improving both segmentation accuracy and interpretability.
2. FFT-based spectral analysis for feature differentiation: Frequency-domain analysis is used to evaluate how preprocessing affects the signal's spectral properties, aiding in feature extraction and wave separation.

To systematically structure this research, the workflow is illustrated in Figure 1, summarizing the step-by-step methodology from data preprocessing to interpretability analysis.

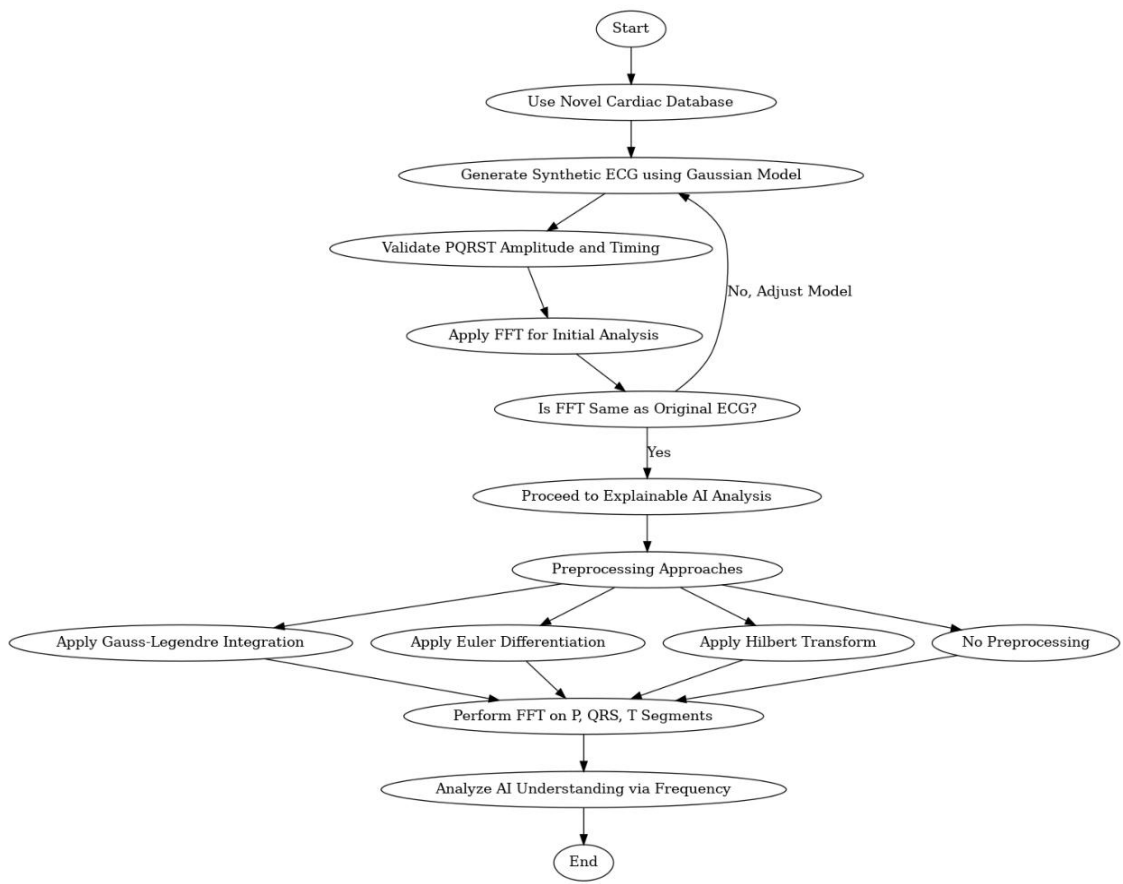

*Figure 1 Research workflow illustrating the process from ECG data selection to Explainable AI-based segmentation. The approach involves synthetic ECG modeling, FFT-based validation, preprocessing (Gauss-Legendre, Euler, or none), and frequency-domain analysis to assess AI interpretability in distinguishing ECG segments.*

By leveraging signal processing principles within an XAI framework, this study provides a transparent and computationally efficient approach to ECG segmentation. The integration of physics-based transformations with deep learning ensures that the model's predictions align with fundamental mathematical properties of ECG signals, enhancing both reliability and practical applicability in real-world biomedical monitoring systems.

## 2 Methods

### 2.1 Dataset

This study employs an ECG dataset introduced in the work titled *"Novel Methodology of Cardiac Health Recognition Based on ECG Signals and Evolutionary-Neural System"* [23], which contains labeled ECG signal fragments corresponding to six distinct cardiac conditions: Normal sinus rhythm, Atrial Premature Beats (APB), Atrial Fibrillation (AF), Atrial Flutter (AFB), Supraventricular Tachycardia (SVT), and Preexcitation syndrome (Wolff–Parkinson–White, WPW). The dataset comprises recordings from 46 individual patients, distributed as follows: 23 with normal rhythm, 9 with APB, 3 with AF, 6 with AFB, 4 with SVT, and 1 with WPW. A total of 538 labeled ECG fragments were extracted, including 283 from normal cases, 66 from APB, 20 from AF, 135 from AFB, 13 from SVT, and 21 from WPW. Each fragment is annotated at the class level, corresponding to the underlying arrhythmia type, providing a suitable foundation for supervised arrhythmia classification. The ECG signals were recorded using clinical-grade equipment with a sampling frequency of 250 Hz, ensuring sufficient temporal resolution to preserve wave morphology. While the original dataset does not include segment-level annotations (e.g., P-wave, QRS complex, and T-wave), additional segmentation was performed in this study using Label Studio, an AI-assisted annotation tool. This segmentation process was manually guided and verified by ECG experts to ensure precise delineation of wave boundaries and to support downstream tasks such as wave-specific learning and interpretable modeling.

### 2.2 ECG Signal Model

Modeling ECG signals as a combination of Gaussian functions provides a precise mathematical representation of the morphology of the P-QRS-T waves. The expression

$$ECG(t) = A_Q \cdot exp\left(-\frac{(t-t_Q)^2}{2 \cdot (\sigma_Q)^2}\right) + A_R \cdot exp\left(-\frac{(t-t_R)^2}{2 \cdot (\sigma_R)^2}\right) + A_S \cdot exp\left(-\frac{(t-t_S)^2}{2 \cdot (\sigma_S)^2}\right) \quad (1)$$

This Gaussian-based approach offers flexibility in capturing the complex dynamics of ECG signals, particularly in cases where morphological variations arise due to physiological or pathological conditions. [3,20,26]. The study by Lim et al. (2024) [8] emphasizes that integrating synthetic models with adaptive segmentation techniques enables more robust feature extraction for ECG classification, especially in detecting arrhythmias with high variability. Furthermore, this approach facilitates the development of automated ECG analysis systems, improving early detection of cardiac abnormalities through machine learning and deep learning techniques. To ensure the synthetic ECG signals closely resemble real physiological signals, we first validate the PQRST morphology using Gaussian-based modeling. Once the amplitude and timing of PQRST components match those in real ECG signals, we apply a Fast Fourier Transform (FFT) analysis to verify spectral similarities. Only synthetic signals with matching spectral characteristics to real ECG data proceed to the Explainable AI analysis. Otherwise, the Gaussian model is adjusted iteratively until spectral alignment is achieved.

### 2.3 Fast Fourier Transform (FFT)

FFT is employed to extract spectral features:

$$X(f) = \sum_{i=0}^{k} x(n) e^{-\frac{j2\pi f n}{N}} \quad (2)$$

These features help differentiate waves based on their frequency components. In synthetic ECG signal analysis using FFT, frequency-domain results highlight unique characteristics of each morphological component, namely the P-wave, QRS complex, and T-wave. FFT results indicate significant differences in frequency distribution among these three segments. Generally, the P-wave, representing atrial depolarization, exhibits dominant energy in the low-frequency range. Meanwhile, the QRS complex, which manifests ventricular depolarization, has a higher dominant frequency, reflecting its rapid and intense nature. The T-wave, associated with ventricular repolarization, once again shows energy in the mid-to-low frequency range, indicating a slower process compared to the QRS complex. This approach makes FFT an effective tool for distinguishing ECG morphology segments based on frequency content. The dominant frequency of each segment can serve as a unique identifier, enabling automatic identification of the P-wave, QRS complex, and T-wave. These findings provide a strong foundation for developing frequency-based ECG signal classification methods, enhancing the accuracy of physiological and pathological heart condition detection. Additionally, this method opens opportunities for integration with machine learning algorithms that utilize frequency-domain data as key features in pattern recognition processes. Thus, FFT-based ECG signal processing not only supports manual analysis but also expands its applications in intelligent diagnostic technology. These visualizations confirm that the frequency characteristics of the synthetic signal closely match those of real ECG signals, validating its suitability for segmentation experiments. Moreover, this further reinforces that each ECG segment—P-wave, QRS complex, and T-wave—exhibits unique frequency characteristics [19,27].

## 2.4 Physics-Based Preprocessing

The Hilbert Transform is utilized to provide an analytic representation of the ECG signal, enabling the extraction of instantaneous amplitude and phase, which are essential for differentiating wave components [17,28,29]. It is mathematically defined as:

$$H(x(t)) = \frac{1}{\pi} P.V. \int_{-\infty}^{\infty} \frac{x(\tau)}{t-\tau} d\tau \qquad (3)$$

where P.V. refers to the Cauchy principal value. This transformation facilitates a clearer distinction between various ECG waveforms. Euler differentiation is employed to emphasize rapid signal changes, especially for detecting sharp transitions like the QRS onset and offset [5,30–32]. The first derivative of the ECG signal is given by:

$$x'(t) = \frac{x(t+\Delta t) - x(t)}{\Delta t} \qquad (4)$$

This method improves sensitivity to sudden changes, helping to better separate ECG segments with rapid transitions. Gauss-Legendre integration is applied to smooth the ECG signal while preserving its morphology [8]. This technique expands waveform structures and enhances spectral separation between ECG segments. The weighted integral is expressed as:

$$\int_{-1}^{1} f(x)dx \approx \sum_{i=1}^{n} w_i \cdot f(x_i) \qquad (5)$$

where $w_i$ weights are based on Legendre polynomials. This integration method refines ECG segmentation by improving feature separation in both time and frequency domains [33–37].

## 2.5 CNN-BiLSTM Model for ECG Segmentation

This study employs a CNN-BiLSTM architecture for ECG signal segmentation, leveraging CNN's spatial feature extraction and BiLSTM's bidirectional temporal pattern capture. LSTM (Long Short-Term Memory) is a type of recurrent neural network (RNN) that excels in recognizing patterns in sequential data [5,6,38–40]. However, traditional LSTMs are limited by their reliance on a single backward propagation process, making it difficult to capture future data context. To address this, the BiLSTM model was developed, allowing access to both forward and backward contextual information. By combining forward and backward LSTMs, BiLSTM enhances feature extraction and improves sequence-based task accuracy.

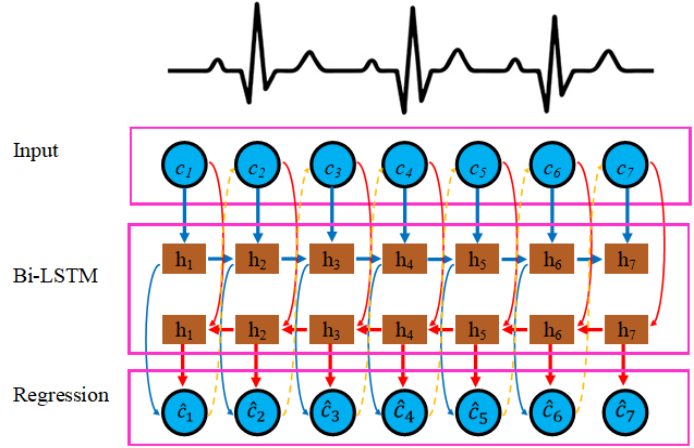

*Figure 2 illustrates the BiLSTM architecture scheme, depicting the flow of data through both the forward and backward layers simultaneously.*

The segmentation process begins with data preprocessing, including downsampling to match the sampling frequency, signal normalization using the mean and standard deviation, and noise removal with a band-pass filter ranging from 0.5 Hz to 50 Hz. Feature extraction is conducted using convolutional operations in CNN, formulated as follows:

$$y(t) = \sum_{i=0}^{N-1} w[i] \cdot x[t+i] + b \tag{6}$$

Where x is the input signal, ww is the filter weight, bb is the bias, and kk is the filter length. The ReLU activation function is applied to introduce non-linearity:

$$f(x) = max(0, x) \tag{7}$$

Pooling layers, such as max-pooling, are used to reduce data dimensionality while retaining essential information.

Next, BiLSTM is utilized to model temporal relationships using input, forget, and output gate mechanisms:

$$i_t = (W_i \cdot [h_{\{t-1\}}, x_t] + b_i) \tag{8}$$

$$f_t = \sigma W_f \cdot [h_{\{t-1\}}, x_t] + b_f \tag{9}$$

$$C_t = f_t \cdot C_{\{t-1\}} + i_t \cdot tanh(W_c \cdot [h_{\{t-1\}}, x_t] + b_c) \tag{10}$$

$$o_t = \sigma(W_o \cdot [h_{\{t-1\}}, x_t] + b_o) \tag{11}$$

$$h_t = o_t \cdot tanh(C_t) \tag{12}$$

The bidirectional approach ensures that temporal information from both forward and backward directions is combined to produce a better signal representation. The final step in this methodology involves segment classification using cross-correlation, which measures pattern similarity between the model's segmented output and reference signals:

$$r_{\{xy\}}(l) = \sum_{\{t=0\}}^{\{N-1\}} x(t) \cdot y(t-l) \tag{13}$$

Where x is the reference signal, y is the predicted result, and ll is the time shift. The correlation value is used to classify segments such as P-wave, QRS complex, and T-wave.

The architecture of the BiLSTM includes layers that perform both forward and backward computations, with hidden states from each direction combined using an activation function to produce the final output. This bidirectional approach enables the model to capture richer context and improve performance on sequential tasks [7,41]. In our study, we extended this approach by using a ConvBiLSTM model, which incorporates convolutional layers to extract spatial features from the data simultaneously. This combination of convolution and bidirectional LSTM effectively processes sequential signals for tasks like segmentation.

*Table 1 The architecture of a Simple ConvBiLSTM Model for ECG PQRST Segmentation. This table outlines the layers and structure of a ConvBiLSTM model used for segmenting ECG signals into P-wave, QRS complex, and T-wave components.*

| Layer (type:depth-idx) | Output Shape | Param # |
| --- | --- | --- |
| ConvBiLSTM | | |
| ├─Conv1d: 1-1 | [1, 64, 3500] | 384 |
| ├─BatchNorm1d: 1-2 | [1, 64, 3500] | 128 |
| ├─Conv1d: 1-3 | [1, 128, 3500] | 24,704 |
| ├─BatchNorm1d: 1-4 | [1, 128, 3500] | 256 |
| ├─Conv1d: 1-5 | [1, 256, 3500] | 98,560 |
| ├─BatchNorm1d: 1-6 | [1, 256, 3500] | 512 |
| ├─LSTM: 1-7 | [1, 3500, 256] | 395,264 |
| ├─Linear: 1-8 | [1, 3500, 2] | 514 |
| Total params: 520,322 | | |
| Trainable params: 520,322 | | |
| Non-trainable params: 0 | | |
| Total mult-adds (Units.GIGABYTES): 1.82 | | |
| Input size (MB): 0.01 | | |
| Forward/backward pass size (MB): 32.31 | | |
| Params size (MB): 2.08 | | |
| Estimated Total Size (MB): 34.41 | | |

Table 1 presents the architecture of the proposed ConvBiLSTM model for PQRST ECG segmentation. The model consists of three Conv1D layers (64, 128, 256 filters), each followed by Batch Normalization to enhance convergence stability. These layers are designed to extract local morphological features from the raw ECG signals. The output is then passed through a bidirectional LSTM to capture temporal dependencies across the full sequence. Finally, a linear layer outputs segmentation logits for each time step. With approximately 520K trainable parameters, this architecture balances segmentation accuracy with computational efficiency and is suitable for resource-constrained environments.

## 3 Results and Discussion

### 3.1 Synthetic ECG Signal Model and Data Visualization

To validate the segmentation model, a synthetic ECG signal was generated based on Gaussian representations of P, QRS, and T waves. The equations used are:

P-wave:

$$P(t) = A_P \cdot exp\left(-\frac{(t-t_p)^2}{2 \cdot (\sigma_P)^2}\right) \quad (14)$$

QRS Complex:

$$P(t) = A_Q \cdot exp\left(-\frac{(t-t_Q)^2}{2\cdot(\sigma_Q)^2}\right) + A_R \cdot exp\left(-\frac{(t-t_R)^2}{2\cdot(\sigma_R)^2}\right) + A_S \cdot exp\left(-\frac{(t-t_S)^2}{2\cdot(\sigma_S)^2}\right) \quad (15)$$

T-wave:

$$T_t = A_t \cdot exp\left(-\frac{(t-t_t)^2}{2\cdot(\sigma_t)^2}\right) \quad (16)$$

Final ECG Signal:

$$ECG(t) = P(t) + QRS(t) + T(t) \quad (17)$$

The Fourier Transform of both synthetic and real ECG signals was computed for frequency domain analysis.

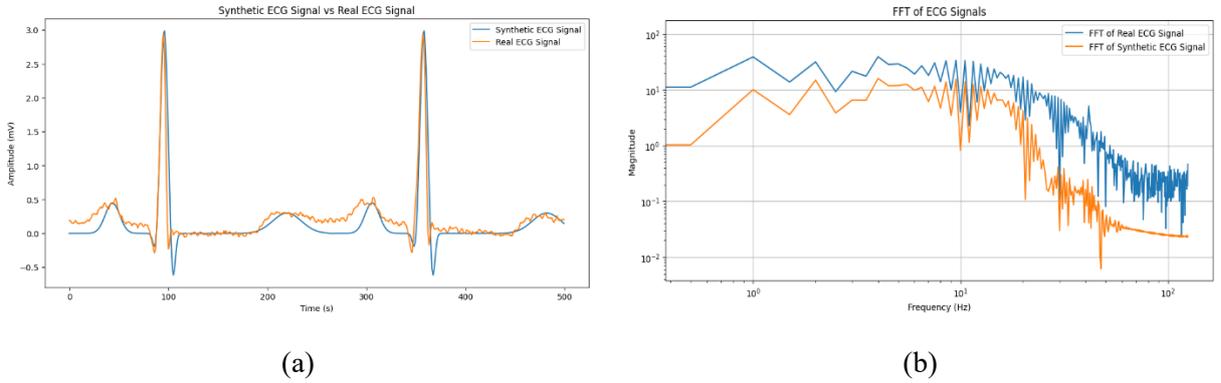

(a)            (b)

Figure 3 *(a)* Comparison of real ECG signal and synthetic ECG signal using the Gaussian-based model. *(b)* FFT comparison between real and synthetic ECG signals.

Unlike many segmentation approaches that rely solely on morphological resemblance, this study emphasizes spectral validation as a prerequisite for trustworthy ECG signal modeling. By employing Fast Fourier Transform (FFT) analysis, synthetic ECG signals are assessed against real clinical recordings to ensure alignment not only in temporal morphology but also in frequency domain characteristics. This dual-layer validation surpasses conventional numeric ECG simulations that often overlook spectral fidelity [3,20,31,42,43]. The model further integrates three preprocessing strategies—Gauss–Legendre integration, Euler differentiation, and raw signal flow—to examine their impact on the separability of P, QRS, and T waveforms based on dominant frequency components. Such comparative evaluation highlights the model's robustness against transformation-induced spectral variations. Previous transform-based methods, including those using FFT, Discrete Cosine Transform (DCT), and Wavelet Transform (WT), primarily target compression or classification objectives without inspecting how signal preprocessing affects spectral segmentation [44,45]. In contrast, Gauss–Legendre schemes have proven suitable for representing oscillatory biomedical signals due to their capacity to preserve waveform integrity in spectral integration tasks [4]. Similarly, Euler-based differentiation enhances steep transitions in ECG morphology, facilitating sharper delineation of QRS boundaries [31,43,46]. By synthesizing frequency-aware preprocessing and spectral benchmarking, this approach offers a transparent, physics-informed foundation for segmenting ECG signals—bridging physiological realism with explainable AI.

## 3.2 ECG Segmentation Performance

The performance of the proposed CNN-BiLSTM model was evaluated by analyzing its ability to segment P-wave, QRS complex, and T-wave from ECG signals. Figure 4 and Figure 5 illustrate the actual and predicted segmentation results, respectively, demonstrating a strong correlation between the ground truth and the model's output. A critical aspect of segmentation evaluation is the model's loss convergence, as shown in Figure 6. The results indicate that the

Gauss-Legendre preprocessing method achieves the lowest training loss, suggesting its effectiveness in enhancing wave-specific feature learning. The raw (unprocessed) signal exhibits the highest loss, reinforcing the importance of preprocessing in improving model performance. The Hilbert and Euler methods show intermediate performance, with the Hilbert transform providing stable convergence and Euler differentiation enhancing sharp transitions.

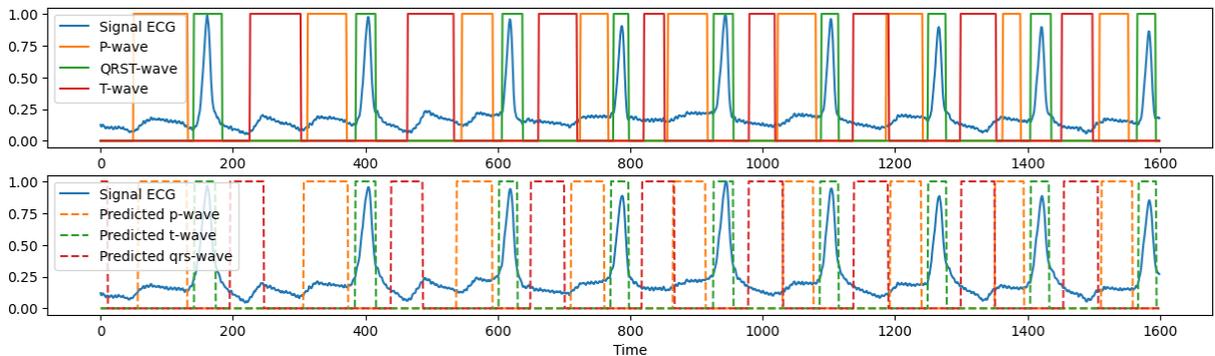

*Figure 4 Actual Segmentation and Predicted Segmentation.*

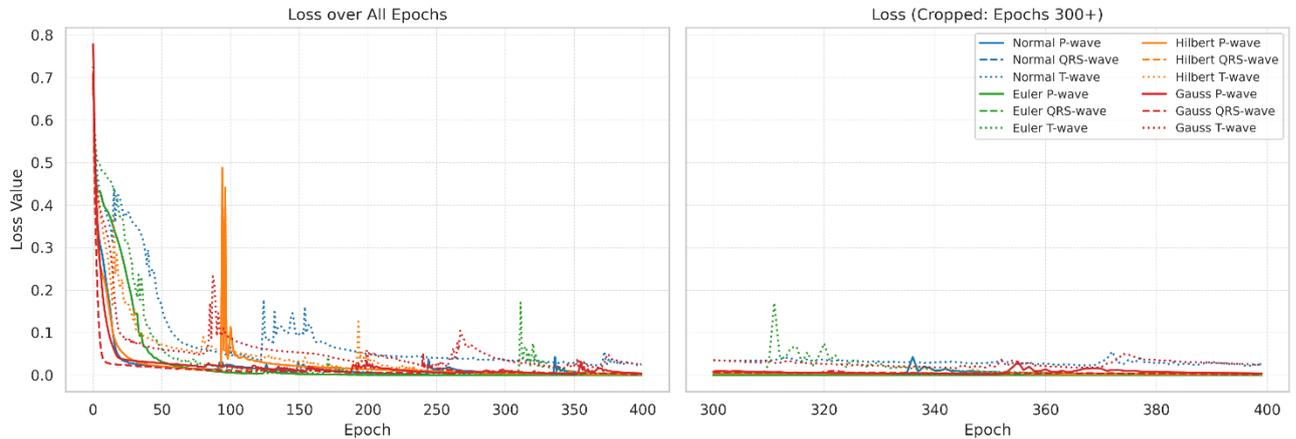

*Figure 5 Training loss comparison for each model using raw (without preprocessing), Hilbert, Gauss-Legendre, and Euler methods, showcasing the progression of loss values over time for P-wave, S-wave, and QRST-wave segmentation. The graph demonstrates how the training loss evolves with each model configuration, revealing the convergence trends and the model's ability to learn wave-specific features effectively. A lower and faster converging loss indicates the effectiveness of different mathematical approaches, with models showing varying performance based on wave complexity and segmentation task.*

As shown in Figure 5, physics-informed transformations yield distinct loss characteristics during model training. The Euler differentiation captures sharp localized transitions, particularly in low-amplitude regions such as the P-wave, enhancing gradient stability throughout training. Gauss–Legendre integration introduces global smoothing effects while preserving wave morphology, leading to steady loss reduction and consistent convergence. The Hilbert transform, through its phase-aware representation, supports morphological alignment across the P-QRS-T complex; however, early training may exhibit ripple-like fluctuations due to complex phase interactions. In contrast, the raw input (without preprocessing) consistently yields higher loss values, especially during late training, reflecting the model's reduced capacity to extract structured temporal features. These outcomes, obtained using a compact CNN–BiLSTM architecture, demonstrate that even lightweight models benefit significantly from carefully designed, physics-based preprocessing. Such transformations not only enhance morphological separability but also embed physiological structure into the learning process. These findings align with prior studies emphasizing preprocessing's role in segmentation and interpretability. For instance, [1,6,7,47] reported performance gains when preprocessing was adapted to cardiac dynamics, while [13,14,48] demonstrated improved multi-lead segmentation through preprocessing-aware modeling strategies. Further validation in [49,50] suggests that preprocessing facilitates convergence and supports the emergence of task-relevant features within deep models. Collectively, these insights affirm that preprocessing is not a peripheral step, but a core design element in high-performance and explainable ECG analysis pipelines.

## 3.3 Implications and Future Directions

The results of this study demonstrate that the integration of spectral analysis and physics-based transformations with deep learning enhances ECG segmentation performance. By leveraging domain-specific signal processing techniques, the model achieves improved accuracy with reduced computational complexity. The explainability provided by the physics-based approach aligns with Explainable AI (XAI) principles, ensuring transparency in ECG analysis. Future research could explore further optimizations, such as incorporating additional adaptive preprocessing techniques or expanding the dataset to improve robustness. The current model's limitations, particularly in T-wave segmentation due to its inherent variability, could be addressed by integrating more sophisticated data augmentation strategies.

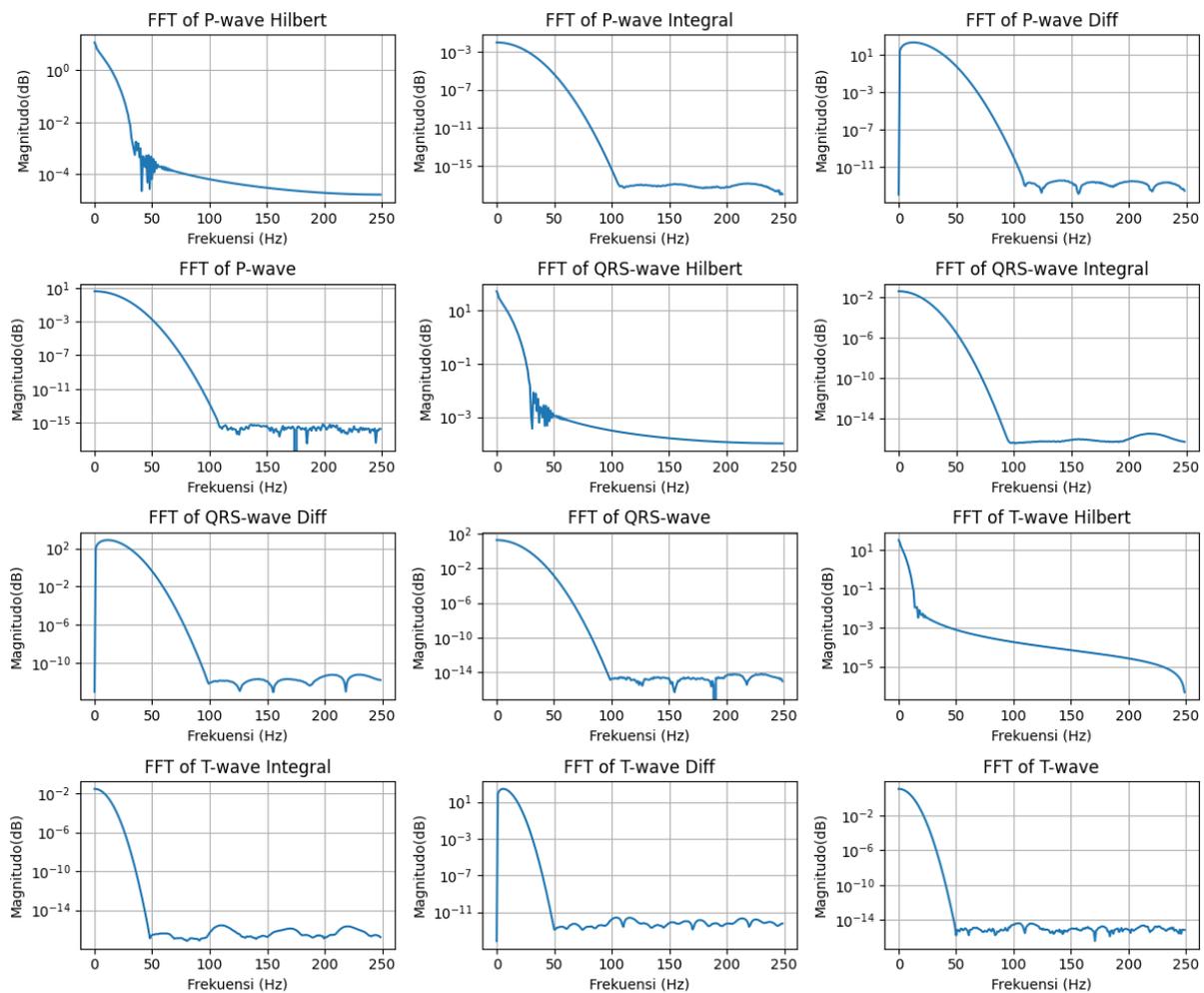

*Figure 6 Frequency domain representation of synthetic signals for each wave segment (P-wave, QRS-wave, and T-wave) using FFT. The analysis compares different preprocessing approaches: raw (without transformation), Hilbert transform, Gauss-Legendre (integration-based smoothing), and Euler (differentiation-based feature enhancement). The spectral characteristics illustrate how each method influences signal representation, highlighting differences in frequency components and potential improvements in feature extraction for ECG segmentation.*

*Table 2 Impact of Preprocessing Methods on ECG Segmentation.*

| Metric | No Preprocessing | Euler Differentiation | Hilbert Transform | Gauss-Legendre Integration |
|---|---|---|---|---|
| Training P-Wave Accuracy (%) | 90.45 | 92.18 | 93.56 | 96.07 |
| Training QRS Accuracy (%) | 95.12 | 96.45 | 96.88 | 97.00 |
| Training T-Wave Accuracy (%) | 89.34 | 91.67 | 92.41 | 93.33 |
| Training Loss | High | Medium | Medium | Low |
| Test P-Wave Accuracy (%) | 89.12 | 90.45 | 92.18 | 96.07 |
| Test QRS Accuracy (%) | 94.50 | 95.12 | 96.45 | 97.00 |
| Test T-Wave Accuracy (%) | 88.00 | 89.34 | 91.67 | 93.33 |
| Test Loss | 0.042 | 0.035 | 0.027 | 0.015 |
| CPU Inference Time (ms) | 40.2 | 41.5 | 43.0 | 45.0 |
| Main Parameter | – | $\Delta t = 0.005$ s | FFT window = 512 | $n = 5$ nodes |
| Advantages | No transformation | Sharp transition detection | Amplitude-phase boost | Accurate & morphologically stable |
| Disadvantages | Noise-sensitive | Less ideal for P/T waves | Risk of waveform shift | Computation-heavy integration |
| p-value vs Raw | – | 0.034 (*) | 0.010 (**) | 0.002 (**) |

Figure 6 shows that the Gauss-Legendre method achieves the lowest training loss, highlighting its ability to preserve signal integrity while enhancing segmentation features. FFT analysis further confirms that Gauss-Legendre significantly refines spectral characteristics across P-wave, QRS-wave, and T-wave segments compared to raw signals. Unlike raw and Hilbert approaches, which may retain overlapping frequencies, Gauss-Legendre expands waveform structures through integration-based smoothing, enhancing segment distinction by amplifying subtle morphological variations and reducing high-frequency noise. This clearer spectral separation improves the model's ability to differentiate wave segments, leading to more precise segmentation and lower training error. The enhanced representation allows the CNN-BiLSTM model to learn wave-specific patterns more efficiently, improving generalization and reducing loss. Table 2 compares preprocessing techniques, demonstrating that Gauss-Legendre achieves the highest segmentation accuracy while maintaining low training loss, making it the most effective method in this study.

## 4 Conclusions

This study presents a lightweight yet effective approach to ECG segmentation by integrating physics-based preprocessing methods with a CNN-BiLSTM model. The use of Gauss-Legendre integration, Hilbert Transform, and Euler Differentiation enhances interpretability and segmentation performance, while FFT analysis provides insights into spectral differentiation between ECG components. The results demonstrate that this approach achieves high segmentation accuracy while maintaining computational efficiency. Compared to traditional deep learning methods, the proposed model reduces reliance on complex architectures and enhances explainability by aligning with well-established signal processing principles. The findings indicate that physics-based transformations can improve model reliability and robustness, making them suitable for real-world biomedical monitoring applications.

Future research should focus on further validating this approach using diverse ECG datasets, optimizing preprocessing techniques, and comparing performance against alternative deep learning models. Additionally, an in-depth analysis of computational efficiency could further support the adoption of this method in clinical and wearable ECG monitoring systems. This study confirms that integrating FFT-based spectral analysis with physics-driven preprocessing enhances ECG segmentation accuracy and interpretability. The ability of the AI model to distinguish ECG wave segments through frequency-domain features highlights the effectiveness of Explainable AI (XAI) in biomedical signal processing. Future research should explore deeper spectral relationships and refine preprocessing strategies to further optimize AI-driven ECG analysis.